\documentclass[10pt,twocolumn,letterpaper]{article}

\usepackage{cvpr}
\usepackage{times}
\usepackage{epsfig}
\usepackage{graphicx}
\usepackage{amsmath}
\usepackage{amssymb}
\usepackage{tabu, multirow}

\usepackage[pagebackref,breaklinks=true,letterpaper=true,colorlinks,bookmarks=false]{hyperref}

\cvprfinalcopy 

\def\cvprPaperID{3559} 

\begin{document}

\title{Deep Chain HDRI: Reconstructing a High Dynamic Range Image from a Single Low Dynamic Range Image}

\author{Siyeong Lee \qquad Gwon Hwan An \qquad Suk-Ju Kang \\
Sogang University\\
Seoul, Republic of Korea\\
{\tt\small \{siyeong, ghan, sjkang\}@sogang.ac.kr}
}

\maketitle

\begin{abstract}
In this paper, we propose a novel deep neural network model that reconstructs a high dynamic range  (HDR) image from a single low dynamic range (LDR) image. The proposed model is based on a convolutional neural network  composed of dilated convolutional layers, and infers LDR images with various exposures and illumination from a single LDR image of the same scene. Then, the final HDR image can be formed by merging these inference results. It is relatively easy for the proposed method to find the mapping between the LDR and an HDR with a different bit depth because of the chaining structure inferring the relationship between the LDR images with brighter (or darker) exposures from a given LDR image. The method not only extends the range, but also has the advantage of restoring the light information of the actual physical world. For the HDR images obtained by the proposed method, the HDR-VDP2 Q score, which is the most popular evaluation metric for HDR images, was 56.36 for a display with a 1920$\times$1200 resolution, which is an improvement of 6 compared with the scores of conventional algorithms. In addition, when comparing the peak signal-to-noise ratio values for tone mapped HDR images generated by the proposed and conventional algorithms, the average value obtained by the proposed algorithm is 30.86 dB, which is 10 dB higher than those obtained by the conventional algorithms.
\end{abstract}

\section{Introduction}
Image restoration is an important field in image processing and computer vision. This task restores an original image using prior knowledge about the degradation phenomena. Unlike image enhancement, the main purpose of image restoration is to restore a latent clean image $\mathbf{x}$ from a corrupted image $\mathbf{y} = H(\mathbf{x})+\mathbf{d}$, where $H$ is the degradation function and $\mathbf{d}$ is additive noise. 

Recently, several approaches for inferring missing information through deep learning have been proposed~\cite{dong2014learning, he2016deep, huang2016densely, kim2016accurate, zhang2017learning}. As a function approximator that infers the unknown mapping between input and output image sets, deep neural networks have advanced state-of-the-art performance in the image restoration field in applications such as super resolution~\cite{dong2014learning, kim2016accurate}, deblurring, and denoising~\cite{zhang2017learning}.

Similarly, efforts to acquire original images close to those in the actual physical world, called high dynamic range imaging (HDRI), have also continued. Debevec \etal~\cite{debevec2008recovering} proposed an HDRI method to expand a narrow (or low) dynamic range due to limitations of the camera sensor. This method estimates the camera response function (CRF) from images with different exposures and derives the radiance information using the estimated CRF, which makes it possible to obtain image information close to the information in the real world. In addition, because of the considerable improvements in display technology, it has been possible to express larger luminance ranges than in the past.  As a result, the interest in generating images with a high quality that is close to that of the real world has increased, and image restoration optimized for high dynamic range (HDR) displays from existing low dynamic range (LDR) images has also become important. Typically, existing images have an LDR that has lost specific information about the captured image due to the camera limitations, and it is impossible to retake these images. Therefore, recovering the dynamic range is an ill-posed problem and can be considered as an image restoration problem. To solve this problem, inverse tone mapping (ITM)~\cite{reinhard2010high} has been proposed. However, conventional ITM algorithms~\cite{akyuz2007hdr,  banterle2009high, huo2014physiological, kovaleski2014high, masia2009evaluation, masia2017dynamic, meylan2006reproduction, rempel2007ldr2hdr} do not infer physical brightness information, but focus on adjusting the brightness values in specific areas such as the highlight regions~\cite{meylan2006reproduction} to create a perceptually high-quality image. In addition, because an image with a narrow range is enlarged to an image with a wide range, it is difficult to find an appropriate relationship between the two spaces with different ranges.

To find the missing information from a single LDR image, we propose a method of restoring an HDR that is close to the real-world range using a deep neural network. The proposed method is a novel HDRI method that produces a multiple exposure image stack from a single LDR image and has the following main contributions.
\begin{enumerate}
\item The neural network expresses the change in the image according to the degree of light exposure, and hence, it produces an HDR image that is close to a real-world image from a single LDR image.
\item We design the neural network with a chain structure to create an LDR image stack by sequentially generating images with different exposure levels from the input LDR image, which is defined as a middle exposure image. In addition, the gradient vanishing problem is solved by inserting an additional loss function so that the learning of the network can be smoothly performed.
\item We propose a new activation function for the HDRI method: Minus PReLU (MPReLU), which transforms the existing PReLU so that residuals between the given input image and an image with a darker exposure are learned easily.
\end{enumerate}

\section{Related work}

\subsection{High dynamic range imaging (HDRI)}
Due to the physical limitations of a charge-coupled device sensor, a digital camera captures a single LDR image with limited dynamic range scene information. This image has a certain exposure value (EV)~\cite{debevec2008recovering}, which is the amount of light that reaches the sensor  of the digital camera. It is controlled by the aperture, shutter speed, and sensor sensitivity, and is defined as follows:
\begin{equation}
EV = 2log_2F - log_2S + log_2(\frac{ISO}{100})
\end{equation}
where $F$ is the relative aperture (F-number), $S$ is the exposure time ($=1/$shutter speed) and $ISO$ is the sensor sensitivity~\cite{ray2000camera}.
When displaying an image with a specific EV, there is a difference between it and the scene observed by a human because the range of the image representation in the camera is smaller than the human perception range. To solve this problem, Debevec \etal~\cite{debevec2008recovering} proposed an HDRI technique that estimates the CRF from multiple LDR images with different exposure levels and extracts an omnidirectional HDR radiance map of the physical world. The relationship between the physical brightness and image pixel level is modeled as follows ~\cite{aguerrebere2012best}:
\begin{equation}
Z_i = f([g_{cv}(C_i + D_i)+N_{reset}]g_{out} + N_{out}) + Q
\end{equation}
where  $f$ is the CRF,  $g$ are the gain of the camera,  $Z_i$ is the pixel value, $C_i$ is the number of photons, $D_i$ is the dark shot noise, $N$ are additional noises and $Q$ is the uniformly distributed quantization error, which occurs during the conversion from analog voltage values to digital quantized values. These methods~\cite{aguerrebere2012best, debevec2008recovering} are limited because it is difficult to capture multiple exposure LDR images for a given scene at the same time. Even if the multiple exposure LDR images are taken and merged to create an HDR image, the method is sensitive to changes caused by moving objects or illumination , thereby degrading the HDR image quality. To solve these problems, several methods have been proposed~\cite{khan2006ghost, min2009histogram,  sen2012robust, srikantha2012ghost} to enhance image quality.

\subsection{Inverse tone mapping (ITM)}
An HDR display has the expanded dynamic range to represent bright and dark areas better than a LDR display. Therefore, displaying the large number of existing LDR images on an HDR display has become a critical issue. For LDR images, there is no information in the saturated regions and dark regions  due to the limitations of the dynamic range. Hence, when an HDR image is generated from a single LDR image, the corresponding regions are difficult to restore. The method for solving the LDR-to-HDR conversion problem is called ITM, and several algorithms have been proposed~\cite{akyuz2007hdr,  banterle2009high, huo2014physiological, kovaleski2014high, masia2009evaluation, masia2017dynamic, meylan2006reproduction, rempel2007ldr2hdr}.
Banterle  \etal~\cite{banterle2009high} proposed an ITM method that detects highlight regions and extends the range of those regions. Masia  \etal~\cite{masia2009evaluation, masia2017dynamic} proposed an exponential expansion method, and Meylan \etal~\cite{meylan2006reproduction} proposed a piecewise linear mapping function that further increases the range for the bright portions of the image. Rempel \etal~\cite{rempel2007ldr2hdr} used a brightness enhancement map to linearly increase the contrast of the intermediate range. Further, they restored the saturated pixel values using an edge stopping function. Kovaleski \etal~\cite{kovaleski2014high} also used a bilateral grid to broaden the dynamic range. However, although these algorithms change an LDR image into an HDR image with a wide range, the expansion~\cite{akyuz2007hdr, masia2009evaluation} is not correctly performed for an image if the parameters are not set appropriately for a given input. Other algorithms~\cite{banterle2009high, kovaleski2014high, rempel2007ldr2hdr} cause contour artifacts on bright objects due to boosting the brightness of the saturated areas, and image quality degrades as a side effect of the additional processing needed to remove annoying artifacts. To solve these problems, Huo \etal~\cite{huo2014physiological} proposed a method that considers the human visual system, using perceptual brightness rather than absolute brightness. However, because it is based on the local adaptive response of the retina, it is difficult to obtain HDR images that match the actual brightness. Recently, Zhang \etal~\cite{zhang2017HDR} proposed a method for converting an LDR panoramic image into an HDR image through deep learning, but the input LDR image has a $64 \times 128 $ pixel resolution, and the method is more suitable for finding the light source position.

\subsection{Convolutional neural network (CNN)}
A CNN automatically extracts a feature map using a loss function defined by the designer when the data set is given and minimizes the error between the inferred value and the reference value. Because of these strong points, CNNs have made many improvements in the field of image restoration. Specifically, in the cases of ResNet~\cite{he2016deep} and DenseNet~\cite{huang2016densely}, it was possible to learn deeper structures through the skip-connections between low-layer information and high-layer abstract information. The VDSR  approach ~\cite{kim2016accurate} obtained good results using residual blocks. Zhang \etal~\cite{zhang2016colorful} also restored colors from grayscale images using CNNs.

\section{Difficulties of direct LDR-to-HDR mapping}
\begin{figure}[t]
\begin{center}
   \includegraphics[width=0.9\linewidth]{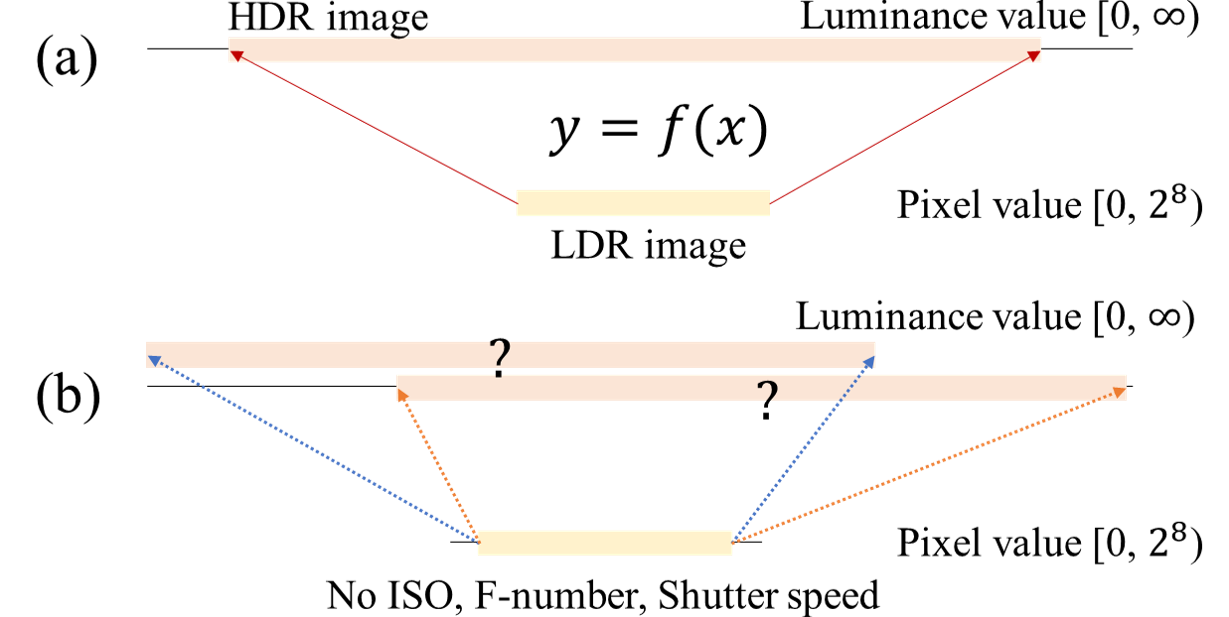}
\end{center}
   \caption{\textbf{ITM problem.} Two problems arise when generating an HDR image from a single LDR image. First, as the range becomes wider, as in (a), there is a lack of mapping information and it becomes difficult to map. Second, if the metadata for the input image does not exist, as in (b), it is impossible to accurately estimate the HDR luminance because the pixel value may be the same depending on the EV, even though it is a different scene.}
\label{fig:ito}
\end{figure}

Before describing details of the proposed neural network architecture, we first explain the feasibility of our neural network structure, which generates a multiple exposure image stack from a single LDR image rather than direct LDR-to-HDR mapping. In terms of restoring the information for an ill-posed problem, a neural network would be an ideal problem solver if it directly extracted the actual scene luminance values from a single LDR image. However, in Figure \ref{fig:ito}(a), as the range required for restoration is widened, it is difficult to infer the relationship between the two image sets. In addition, it is impossible to simply expand the dynamic range, as shown in Figure \ref{fig:ito}(b) and hence  the metadata (e.g., sensor sensitivity, F-number, and shutter speed) of the input LDR image are required to infer the luminance values  of the actual scene. Generally, most existing LDR images do not have EV information, and hence, there is a distinct limitation to restoring the actual scene luminance values. We assume that the existing LDR images are well-captured or properly exposed because the appropriate (or optimal) exposure value will have been selected by humans. The single input LDR image of the proposed network is defined as a middle exposure (EV 0) image. (A description of middle exposure in the proposed neural network is given in Section \ref{sec:dataset}.) By assuming a middle exposure and using a multiple exposure image stack, it is possible to train and infer $\pm 1, \pm 2$, and $\pm 3$  EV LDR images that contain higher or lower exposure information, as shown in Figure 2. Even if the EV value is not known for the existing LDR image, we can obtain a tone mapped HDR image that is well-fitted to the wider range display. In addition, if the EV for the input LDR image is known, the scene luminance of the HDR can be inferred through Debevec \etal's method~\cite{debevec2008recovering}.
Therefore, we propose a neural network that infers the multiple exposure image stack from a single LDR image to find the relationship between the HDR image from the LDR image, which is defined as a middle exposure image.
\begin{figure}[t]
\begin{center}
   \includegraphics[width=0.85\linewidth]{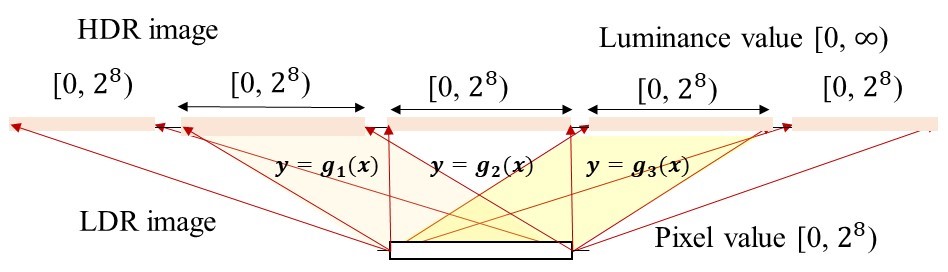}
\end{center}
   \caption{\textbf{Proposed multiple exposure image stack.} To obtain an HDR image using the proposed method, several subnetworks generate LDR images with various exposure levels rather than inferring the entire part at once.}
\label{fig:proposed}
\end{figure}
\begin{figure*}[t]
\begin{center}
   \includegraphics[width=0.85\linewidth]{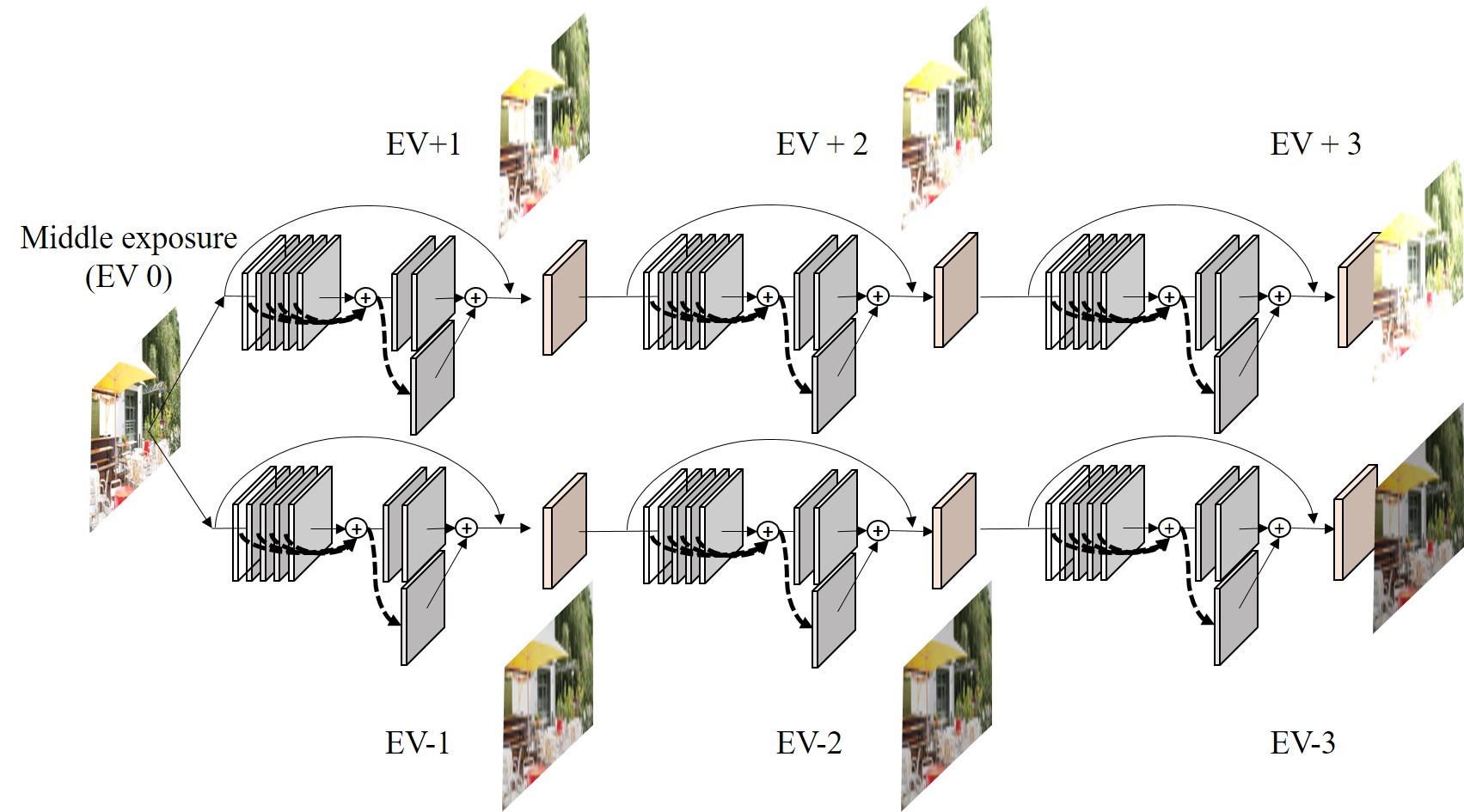}
\end{center}
   \caption{\textbf{Proposed Deep Chain HDRI architecture.} Given an LDR image with a middle exposure value (EV $0$), the EV $\pm 1, \pm2, \pm 3$  images are inferred sequentially through the network. When the EV of the inferred image is far from the middle exposure value, the structure depth goes deeper than that of the image that has less exposure difference to infer the mapping relation more accurately. After finishing the process through the proposed network, a total of six LDR images are inferred to generate the LDR image stack. Then, an HDR image is generated using the HDRI technique.}
\label{fig:overall}
\end{figure*}

\section{Proposed Deep Chain HDRI}
\subsection{Overall network architecture}
Based on an LDR image with a middle exposure, the proposed neural network has the structure shown in Figure \ref{fig:overall}. The proposed network, which consists of six subnetworks, infers images with different exposures that are higher (or lower) than that of the input image. (This is discussed in detail in Section \ref{sec:chain}.) When the exposure level is further from the middle exposure, a deeper structure is needed to infer the relationship between the input and output images. Therefore, the proposed network constructs the sequential learning process, and it is then possible to increase the depth of the neural network to infer between patches that are far from the given exposure information. The entire network produces images with the top three exposures and the bottom three exposures from the input LDR image with the middle exposure.

\subsection{Subnetwork architecture}
\begin{figure}[t]
\begin{center}
   \includegraphics[width=0.95\linewidth]{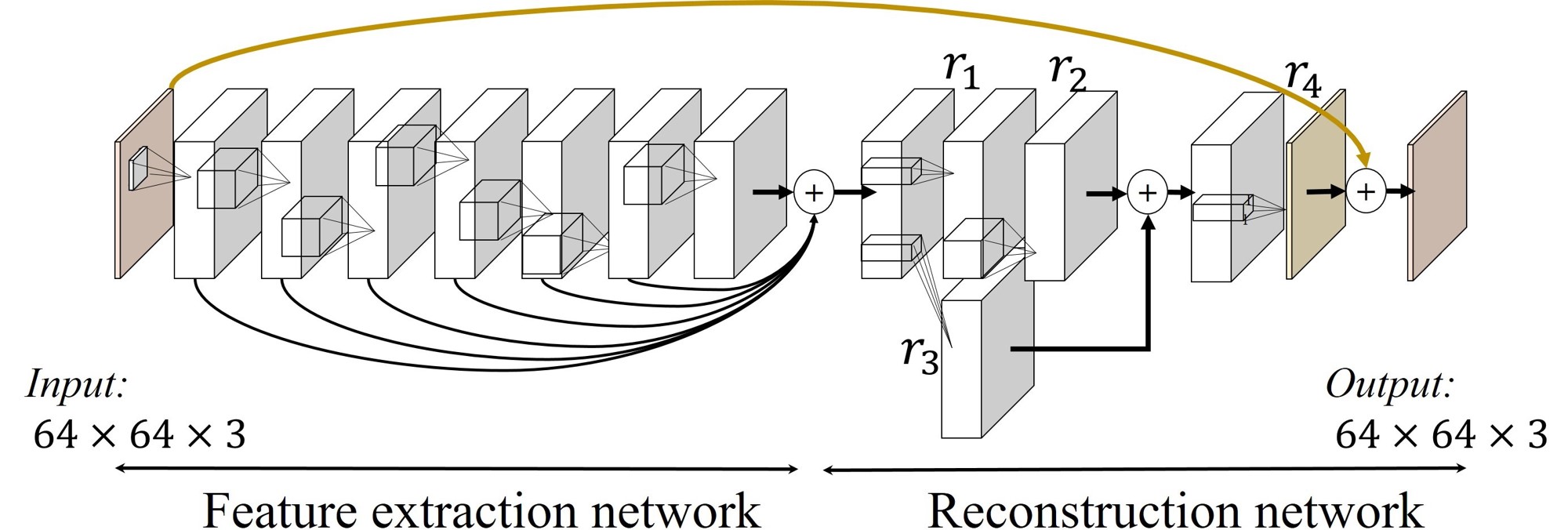}
\end{center}
   \caption{Subnetwork architecture.}
   \label{fig:sub}
\end{figure}

As shown in Figure \ref{fig:sub} , we use a CNN-based subnetwork to create an LDR image from a given EV $i$ to EV $j$. The subnetwork is affected by the DCSCN  architecture~\cite{yamanaka2017fast}. It uses a $64 \times 64$ LDR patch in the image with an EV of $i$ as input and produces a $64 \times 64$ LDR patch with an EV of $j$ as the output. The subnetwork is divided into two parts. The front part consists of a total of seven feature extraction blocks, and the rear part consists of four reconstruction blocks. Each block consists of a (dilated) convolution layer, a batch normalization layer, and an activation layer. We use PReLU~\cite{he2015delving} as the activation function for a block that infers brighter images and MPReLU, which is first proposed in this paper, as an activation function for a block that infers darker images. (Additional description of MPReLU is given in Section \ref{sec:residual}.) The feature extraction network operates as an abstraction of feature extraction for a given input. The reconstruction network consists of $r_1,  r_2,  r_3$, and $r_4$  with two paths, and reconstructs the image using the extracted features, as shown in Figure \ref{fig:sub}. Unlike the layers of other reconstruction networks, $r_4$ uses the \textit{tanh} function as an activation function to enable the representation of the residual. Additionally, when extracting feature maps from a CNN, the receptive field is significant. Therefore, we set the dilation parameters of the convolution layer of the feature extraction block to $1, 1, 2, 3, 5, 8$, and $13$, respectively, so that we can take into account all the information in the patch. Hence, the receptive field of this network is $67$. Each layer in the feature extraction network has $32$ kernels in $3\times3$ size. In the reconstruction network, each layer consists of $32$ kernels. The kernel size of $r_1,  r_2,  r_3$ and $r_4$ are  $ 1\times1, 1\times1 , 3\times3$, and $1\times 1$, respectively.

\subsection{Training}
To train the proposed model, the input of subnetwork $N_j$, which estimates the image with EV $j$ from EV $i$, maps between the ground truth image with EV $i$ (or the image inferred from the previous subnetwork) and the image with the ground truth image with EV $j$. The optimization proceeds in the direction of minimizing these losses. Each subnetwork uses batch normalization to produce regularization effects and prevent the vanishing gradients that can occur during learning. In addition, by adding an error of EV  $\pm 1$ and $\pm 2$  between the subnetwork structures, vanishing gradients, which can occur due to the deep structure for EV $\pm 3$  learning, are prevented.
\\

\noindent
\textbf{Loss function} 
The proposed CNN structure is a network structure with multiple outputs from one input image.  Given the output  $\hat{y}_i$ of EV $i$ and the target image $y_i$, we set the loss function as follows.
\begin{equation}\label{eq:totalloss}
\mathbb{L}_{all}(\hat{y_i}, y_i) = l_{pixel}(\hat{y_i}, y_i) + \lambda l_{tv}(\hat{y_i})
\end{equation}
where $ l_{pixel}$ is the pixel loss and  $ l_{tv}$ is the total variation regularization. To improve the smoothness of the inferred image and prevent it from overfitting to a specific pattern, we add total variation regularization. Experimentally, we set the relative weights of each loss to $\lambda = 0.001$. Therefore, the entire network trains to minimize the loss of each of  $\hat{y}_i$ and $y_i$. The pixel loss is an L1 norm between $\hat{y}_i$ and $y_i$. 
\begin{equation}
 l_{pixel}(\hat{y_i}, y_i) = \frac{1}{wh}\sum_{u,v}||\hat{y_i}(u,v) - y_i(u,v)||_1
\end{equation}
where $w$ is the width of the image, $h$ is the height of the image, and $u$ and $v$ are the pixel coordinates.
\\

\noindent
\textbf{Patch-based learning} The brightness caused by light radiation in an image has a localized property. Therefore, we make a set of $64 \times 64$  patches $p \in [0,255]^{64\times64}$ from the image $I \in [0,255]^{w \times h}$  corresponding to EV $i$ with a stride of $10$. As a result, each subnetwork that infers an EV $j$ image from an EV $i$ image is trained with about $180,000$ patch pairs.
\\

\noindent
\textbf{Residual learning} 
In the case of image transformations that learn the relationship between different images, a neural network often loses the morphological information from a given input image while optimizing the loss minimization. Therefore, to avoid losing spatial information, the neural network is designed to learn the residual image from ground truth. The batch size was set to one. To optimize the weights and biases of the neural network, we used the Adam optimizer~\cite{kingma2014adam} with a learning rate of $0.001$ and momentum parameter $\beta_1$  of $0.9$. In addition, each subnetwork is trained using a \textit{Nvidia GeForce 1080 Ti GPU} with $100$ epochs for about $48$ hours.

\subsection{Dataset}
\label{sec:dataset}
\noindent
\textbf{Middle exposure} The correct exposure is difficult to define because it reflects a subjective viewpoint. However, we assume that a correctly exposed image depicts all parts of the image in detail. In other words, pixels are uniformly distributed throughout the grayscale range. This is defined as a middle exposure from a technical point of view. Therefore, when a multiple exposure image stack is obtained by the auto bracketing function of the camera, which changes the EV automatically when capturing images, we can determine the middle exposure image, which is the image with the most evenly distributed histogram of the images of the stack. In addition, we define the EV of the corresponding image as the middle exposure value (EV $0$).
\\

\begin{figure}[h]
\begin{center}
   \includegraphics[width=0.95\linewidth]{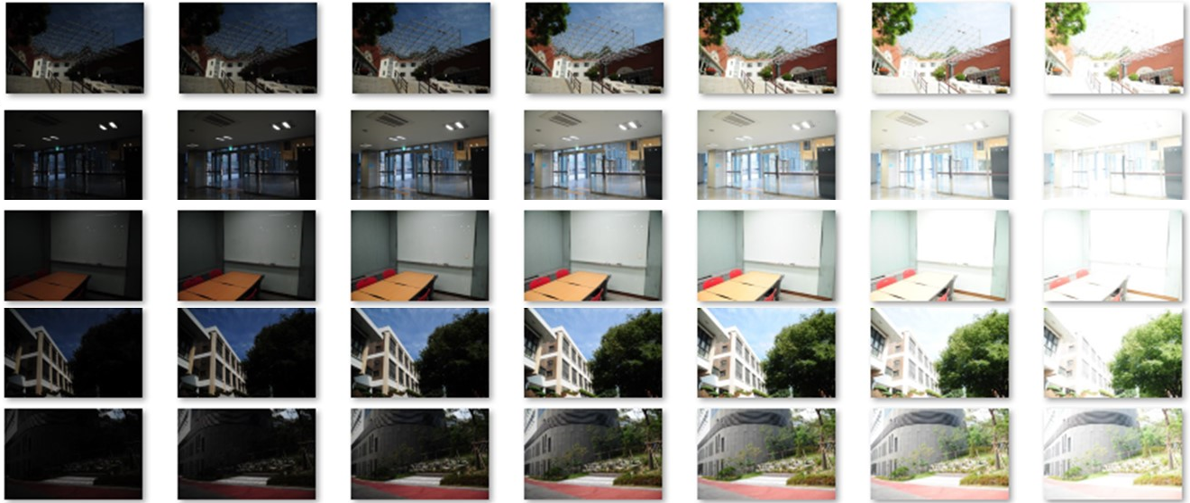}
 \end{center}
  \caption{Sample images for the new dataset consisting of multiple exposure image stacks.}
  \label{fig:dataset}
\end{figure}

\noindent
\textbf{New data set of multiple exposure image stack} For learning each subnetwork, we need seven multiple exposure ground truth images satisfying EV $0, \pm 1, \pm 2, \pm 3$ for static scenes. Generally, many datasets are required to train a neural network, but only five stacks satisfy this condition among the existing HDR datasets~\cite{EMPA, HDRI}. Therefore, we generated $96$ different scene image stacks ($672$ images; outdoor: $504$, indoor: $168$ images) to train and test the proposed network. We shuffled and split the dataset into a training set, validation set, and test set. The ratio of each set is $7:3:10$, respectively. We captured all the images in an $8$ $bit$ JPG format using a Nikon D700 with a resolution of $4256 \times 2832$, which was resized to  $912 \times 608$  to obtain the appropriate training rate. We used a tripod to minimize image blur, set the aperture value to $f/4$, and automatically adjusted the sensor sensitivity and shutter speed using the auto bracketing function. Then, EV $\pm 1, \pm2, \pm3$ images from the middle exposure image were stored, as shown in Figure \ref{fig:dataset}.

\subsection{HDRI reconstruction}
Given image $I \in [0,255]^{w \times h}$ with EV $0$, we sliced $I$ into the  $64 \times 64$  LDR patches with a stride of $1$. After transforming the image into a tensor shape, we inferred a patch with the upper (or lower) EV through the neural network and reconstructed $I_{ev\pm1} \in [0,255]^{w \times h}$ using the reconstruction process. Then, the process was repeated to generate LDR images with other EVs and construct the multiple LDR image stack. After that, an HDR image was merged using the HDRI method proposed by Debevec \etal~\cite{debevec2008recovering}.

\section{Understanding the properties of the proposed method}
In this section, we analyze the reasons for the validity of the proposed neural network architecture and the characteristics of the network. First, we examine the necessity for having a chain structure as the exposure difference increases. Second, we analyze the settings of the activation function and its properties to solve the problems that may arise through residual learning for various exposures in the proposed method.

\begin{table}[b]
\begin{tabu} to \linewidth {X[c]X[c]X[c]X[c]X[c]X[c]X[c]}
\tabucline[1.2pt]{-}
\rowfont{\scriptsize \bfseries}
EV -3 &  EV -2 &  EV -1  & EV 0 &  EV +1  &  EV +2 &  EV +3 \\
\tabucline[0.7pt]{-}
\rowfont{\small}
8.7        & 10.87     & 15.74      &  $\infty$ & 15.88       & 10.40      & 7.62 \\
\tabucline[1.2pt]{-}
\end{tabu}
\caption{Average PSNR between the middle exposure image and various EV images.}\label{tab:distance}
\end{table}

\subsection{Why need the deep chain structure?}
\label{sec:chain}
The average peak signal-to-noise ratios (PSNRs) between the ground truth images with different EVs of the same scene are calculated in Table \ref{tab:distance}. This difference arises because the information in  the amount of light entering the camera gradually changes as the EV is changed. When the EV difference is relatively large, the two images are further apart. It can be assumed that a relatively deeper neural network is required when inferring the relationship between two such images. Hence, we designed a neural network structure not just simply by increasing the depth structure, but by also proposing a chain structure, which infers the EV sequentially. For example, EV $\pm3$ images can be made after inferring EV $\pm1$ and $\pm2$ images sequentially from an input image (EV $0$). 

To validate the chain structure, we compared the proposed method with a relatively shallow network that is skip-connected between three convolution layers and three deconvolution layers~\cite{mao2016image} for the problem of inferring the relationship between EV $0$ and EV $+3$. The results are shown in Table \ref{tab:skipped}. The overall result shows that when inferring the relationship between images with a large distance, the deeply structured neural network is better. Therefore, it can be concluded that it is good to have a deeper structure when the exposure difference increases between the images. In addition, as the depth of the structure increases, the gradient vanishing effect, where the error cannot be delivered to the end during backpropagation, may occur. Therefore, a loss term is added to the intermediate results, which correspond to the EV $\pm1$ and  $\pm2$ images during the process. As a result, the proposed neural network architecture infers EV $\pm3$ images more accurately.

\subsection{Residual learning and activation function}
\label{sec:residual}
In the deep neural network structure, neuron activation is determined by a nonlinear function to describe the nonlinear relation between the input and output. Based on the output of each neuron, the \textit{sigmoid} and ReLU~\cite{maas2013rectifier} functions have a non-negative output. In contrast, ELU~\cite{clevert2015fast} and SELU~\cite{klambauer2017self} have a gradual slope toward negative infinity, and PReLU~\cite{he2015delving} changes the slope in the negative domain. The proposed neural network contains a residual learning process that learns the difference between the input and reference. For the residuals of the image with a lower EV than the input LDR image, images with  EVs of $-1, -2, -3$ decrease in pixel value. This means that the weight and bias values become more negative. Because of this perspective, when the EV of the image decreases, it is difficult to find a relation with functions such as the ReLU function. Therefore, the proposed neural network requires an activation function that can reflect a negative value such as PReLU. However, in the negative domain of PReLU, the error does not flow easily due to the gradient variation, and it has the relatively smaller slope than in the positive domain, which flows consistently with the backpropagation.

Accordingly, we propose a new activation function for the HDRI method: MPReLU, which is an extension of the existing PReLU. It is defined as follows.
\begin{equation}
\textrm{minus PReLU}(x) = \left\{ \begin{array}{ll}
\alpha x & \textrm{if $x \geq 0$}\\
x & \textrm{if $x < 0$}
\end{array} \right.
\end{equation}
The comparison results are shown in Table \ref{tab:mprelu}. These results show that MPReLU is effective for learning the residuals between a given input image and an image with a darker exposure. Hence, in the proposed neural network architecture, PReLU is used for images with a higher exposure (brighter images), and MPReLU is used for images with a lower exposure (darker images) to train the network.

\begin{table}[t]
\begin{tabu} to \linewidth {X[2,c]X[1,c]X[1,c]X[1,c]X[1,c]X[1,c]X[1,c]}
\tabucline[1.2pt]{-}
\multirow{2}{*}{\textbf{Method}} & \multicolumn{2}{c}{\textbf{PSNR(dB)}} & \multicolumn{2}{c}{\textbf{SSIM}} & \multicolumn{2}{c}{\textbf{MS-SSIM}} \\
\tabucline[0.7pt]{2-7}
                  & $m$        & $\sigma$        & $m$      & $\sigma$      & $m$       & $\sigma$      \\ 
\tabucline[0.7pt]{-}
\textit{Ours}              &  28.18 &  2.77 & 0.953 & 0.065 & 0.983 & 0.015 \\
\tabucline[0.7pt]{-}
{[}22{]}        &  17.52 &  4.83 & 0.913 & 0.077 & 0.966 & 0.048\\ 
\tabucline[1.2pt]{-}
 \end{tabu}
\caption{Validity of the chain structure neural network.}\label{tab:skipped}
\end{table}

\begin{table}[t]
\begin{tabu} to \linewidth {X[1,c]X[2,c]X[2,c]X[2,c]X[2,c]X[2,c]X[2,c]X[2,c]}
\tabucline[1.2pt]{-}
\multicolumn{2}{c}{\multirow{2}{*}{}} & \multicolumn{2}{c}{\textbf{PSNR(dB)}} & \multicolumn{2}{c}{\textbf{SSIM}}    & \multicolumn{2}{c}{\textbf{MS-SSIM}}   \\ \tabucline[0.7pt]{3-9}
\multicolumn{2}{c}{}                  & $m$               & $\sigma$ & $m$              & $\sigma$ & $m$              & $\sigma$ \\ \tabucline[0.7pt]{-}
\multirow{2}{*}{\scriptsize EV -1}    & \scriptsize MPReLU    & \textbf{29.01}  & 3.83     & \textbf{0.935} & 0.056    & \textbf{0.980} & 0.017    \\
                          & \scriptsize PReLU     & 28.28             & 2.96     & 0.931            & 0.053    & 0.977            & 0.015    \\ \tabucline[0.7pt]{-}
\multirow{2}{*}{\scriptsize EV -2}    & \scriptsize MPReLU    & \textbf{26.72}  & 4.54     & \textbf{0.952} & 0.029    & \textbf{0.974} & 0.021    \\
                          & \scriptsize PReLU     & 24.98             & 3.92     & 0.910            & 0.050    & 0.962            & 0.028   \\ \tabucline[0.7pt]{-}
\multirow{2}{*}{\scriptsize EV -3}    & \scriptsize MPReLU    & \textbf{24.33}  & 4.57     & \textbf{0.919} & 0.036    & \textbf{0.948} & 0.037    \\
                          & \scriptsize PReLU     & 22.58             & 4.46     & 0.836            & 0.075    & 0.933            & 0.046    \\ \tabucline[1.2pt]{-}
 \end{tabu}
\caption{Comparison of MPReLU and PReLU.}\label{tab:mprelu}
\end{table}

\begin{figure*}[h]
\begin{center}
   \includegraphics[width=0.96\linewidth]{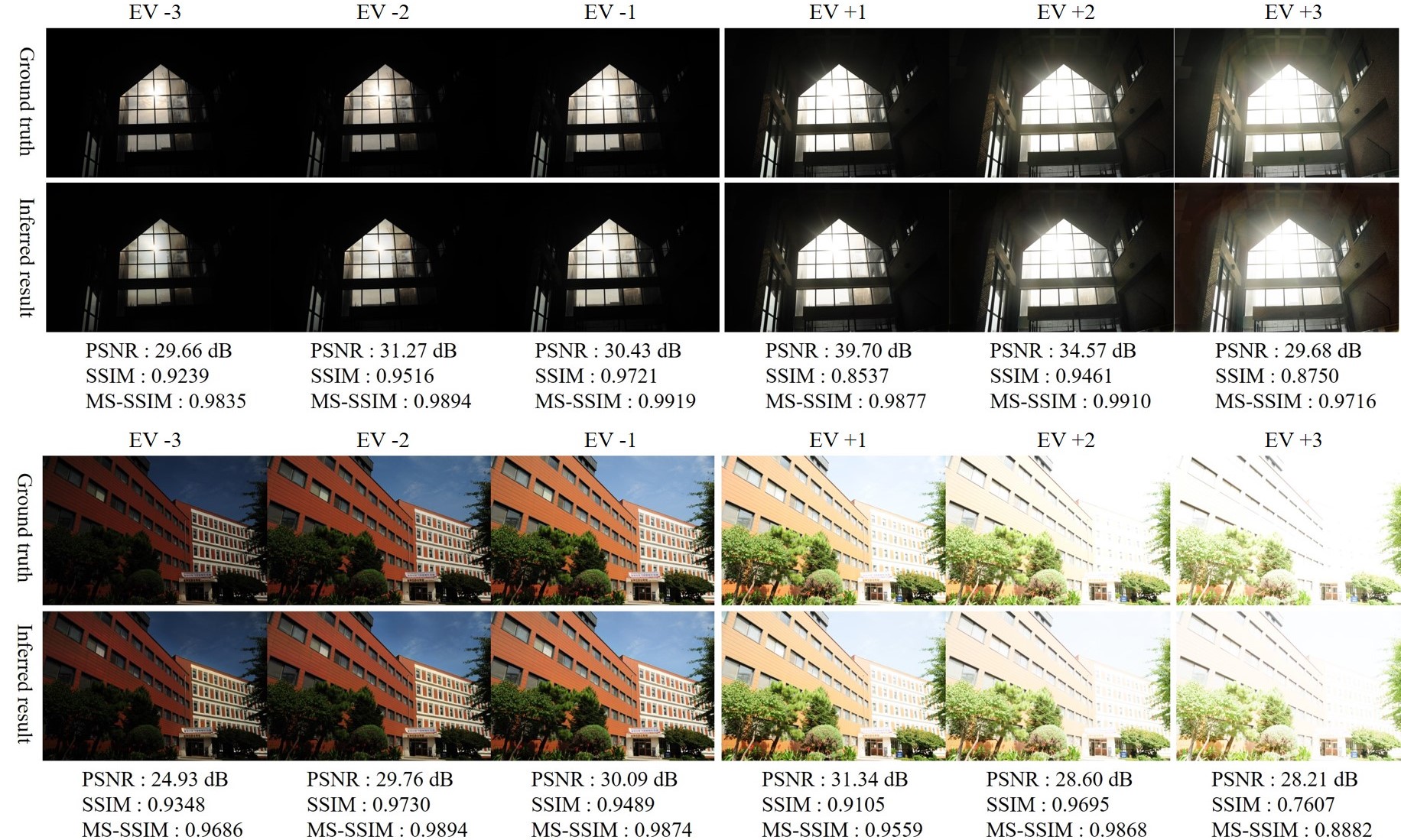}
\end{center}
   \caption{\textbf{Comparison of the ground truth LDR image stack and inferred LDR image stack.} The proposed neural network follows not only the actual variation trends of the exposure, but also the properties of the light source.}\label{fig:LDR-HDR}
\end{figure*}

\section{Experimental results}
The results of the proposed network are divided into two parts: (1) a comparison between the ground truth LDR image stack and the inferred LDR image stack and (2) a comparison between the ground truth HDR images and inferred HDR images. The conventional ITM algorithms~\cite{huo2014physiological, kovaleski2014high, masia2017dynamic} were used. In the experiments, $48$ image stacks were used for the comparisons, and the HDR Toolbox~\cite{banterle2017advanced} was used to generate, tone map, and compare the HDR images with those obtained by the conventional ITM algorithms.

\begin{table}[b]
\begin{tabu} to \linewidth {X[2,c]X[1,c]X[1,c]X[1,c]X[1,c]X[1,c]X[1,c]}
\tabucline[1.2pt]{-}
\multirow{2}{*}{\textbf{EV}} & \multicolumn{2}{c}{\textbf{PSNR(dB)}} & \multicolumn{2}{c}{\textbf{SSIM}} & \multicolumn{2}{c}{\textbf{MS-SSIM}} \\
\tabucline[0.7pt]{2-7}
                  & $m$        & $\sigma$        & $m$      & $\sigma$      & $m$       & $\sigma$      \\ 
\tabucline[0.7pt]{-}
EV +3              &  28.18 &  2.77 & 0.953 & 0.065 & 0.983 & 0.015 \\
\tabucline[0.7pt]{-}
EV +2              &  29.65 &  3.06 & 0.959 & 0.065 & 0.986 & 0.016 \\
\tabucline[0.7pt]{-}
EV +1              &  31.90 &  3.43 & 0.969 & 0.039 & 0.992 & 0.008 \\
\tabucline[0.7pt]{-}
EV -1              &  29.01 &  3.83 & 0.935 & 0.056 & 0.980 & 0.017 \\
\tabucline[0.7pt]{-}
EV -2              &  26.72 &  4.54 & 0.952 & 0.029 & 0.974 & 0.021 \\
\tabucline[0.7pt]{-}
EV -3              &  24.33 &  4.57 & 0.919 & 0.036 & 0.948 & 0.037 \\
\tabucline[1.2pt]{-}
 \end{tabu}
\caption{Comparison of the ground truth LDR image stack and inferred LDR image stack.}\label{tab:exp}
\end{table}

\subsection{Comparison between the ground truth LDR image stack and inferred LDR image stack}
To determine the similarity between the inferred LDR images and the ground truth LDR images, PSNR, structural similarity (SSIM), and multi-scale SSIM (MS-SSIM) were used. The comparison results are shown in Table \ref{tab:exp} and Figure \ref{fig:LDR-HDR}. The similarity between the inferred LDR images and the ground truth images decreases as the exposure difference from EV $0$ increases. In addition, the images with a brightness that is darker than that of the single input LDR image are less similar to the ground truth than the images with a brightness that is brighter than the input single LDR image. It is assumed that it is more difficult to infer relative darkness from an input LDR image.

\begin{figure*}[t]
\begin{center}
   \includegraphics[width=0.97\linewidth]{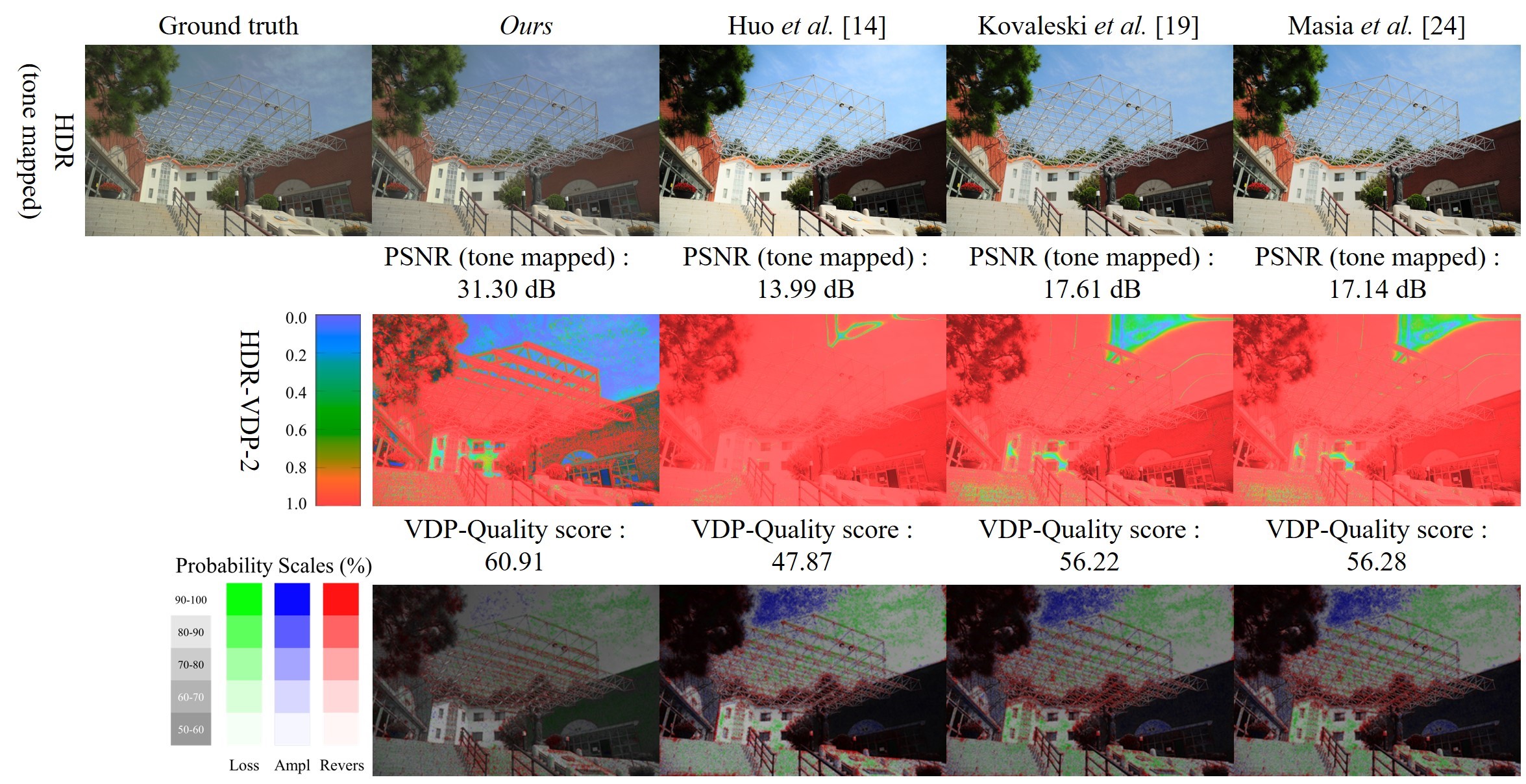}
\end{center}
  \caption{Comparison of the ground truth HDR image with HDR images inferred by the proposed and conventional methods.} \label{fig:HDR-HDR}
\end{figure*}

\subsection{Comparison between ground truth HDR images and inferred HDR images}
HDR images were generated using the ground truth LDR image stack and the inferred LDR image stack. In the process of HDRI with the LDR image stacks, the CRF was estimated based on Debevec \etal~\cite{debevec2008recovering}. To tone-map the HDR images, a representative tone mapper, Reinhard\etal's method~\cite{reinhard2002photographic}, was used. 
The PSNR among the tone-mapped HDR images was used for a quantitative comparison of the HDR images. The HDR images were evaluated through HDR-VDP-2~\cite{mantiuk2011hdr} and DRIM~\cite{aydin2008dynamic}, which are based on the human cognitive system. The evaluation used the input parameters of a $24''$ display, $0.5$ m, viewing distance, $0.0025$ peak contrast, and $2.2$ gamma. The evaluation results are shown in Table \ref{tab:final} and Figure \ref{fig:HDR-HDR}. The PSNR among the tone mapped HDR images and the VDP-Quality score among the HDR images quantitatively show how much more closely to the ground truth image the proposed method infers than the other methods. In the conventional methods, it is difficult to infer the actual scene luminance because the single LDR image information is simply expanded to fit the target dynamic range. Therefore, there are artifacts such as brightness boosting and loss of details in some areas. In addition, the HDR-VDP-2 and DRIM results also show that the proposed method approximates the actual scene luminance better than the conventional algorithms. In the result image of HDR-VDP-2, the proposed method had more blue-color pixels than the existing methods. A result image with colors close to $0$ (blue) means that the observer cannot recognize the difference from the ground truth HDR image. The DRIM shows the contrast reversal, loss of visible contrast, and amplification of invisible contrast through red, green, and blue points, respectively, to represent the differences between HDR images. Because the proposed method aims to infer the actual scene luminances, it can be seen that the best-inferred result is when there are no red, green, and blue points in the DRIM result image. From this point of view, the proposed neural network inferred the actual scene luminance from a single LDR image better than the conventional methods.
\begin{table}[h]
\begin{tabu} to \linewidth {X[2,c]X[1,c]X[1,c]X[1,c]X[1,c]}
\tabucline[1.2pt]{-}
\multirow{2}{*}{\textbf{Method}} & \multicolumn{2}{c}{\begin{tabular}[c]{@{}c@{}}\textbf{PSNR(dB)}\\\small Tone mapped\end{tabular}} & \multicolumn{2}{c}{\begin{tabular}[c]{@{}c@{}}\textbf{VDP-Quality}\\\textbf{score}\end{tabular}} \\
\tabucline[0.7pt]{2-5}
                   & $m$               & $\sigma$ & $m$ & $\sigma$ \\ 
\tabucline[0.7pt]{-}
\textit{Ours}             &  30.86 & 3.36 & 56.36 & 4.41 \\
\tabucline[0.7pt]{-}
{[}14{]}      &  18.43 &  3.04 & 50.00 & 5.86 \\
\tabucline[0.7pt]{-}
{[}19{]}      &  21.76 &  2.81 & 50.28 & 4.98 \\
\tabucline[0.7pt]{-}
{[}24{]}      &  20.13 &  2.21 & 51.24 & 5.67 \\
\tabucline[1.2pt]{-}
 \end{tabu}
\caption{Comparison of the ground truth LDR image stack and inferred LDR image stack.}\label{tab:final}
\end{table}

\section{Conclusion}
In this paper, we proposed a novel artificial neural network structure that infers an HDR image from a single LDR image. By inferring the suitable image luminance for the scene when the given LDR image is captured, the proposed network structure not only widens the dynamic range of the LDR image, but also generates an image that is closer to the ground truth image than the previously proposed methods. Moreover, the proposed network trains the residuals in the image pair, which contains the morphological information and changes in illumination, from the given training set. The proposed subnetwork serves as a dictionary that contains the brightness information for each image with the desired exposure level by rearranging the images sequentially in the exposure space. 

As a result, the proposed network is able to solve problems such as ghosting and tearing, which appear in conventional HDRI. Furthermore, the proposed network is scalable in that it can be further extended to obtain a far wider dynamic range. In addition, because patch-based learning has been carried out, it is less restricted to the image resolution for restoring HDR images. As shown in the experimental results, learning relatively low exposures is difficult. The further improvement and additional studies about the network structure will be addressed in future work.\\

\newpage

{\small
\bibliographystyle{ieee}
\bibliography{egbib}
}

\end{document}